\documentclass{llncs} %

\usepackage{float}
\usepackage{graphicx}
\usepackage{wrapfig}

\usepackage[labelfont=bf]{caption}
\usepackage{color,soul}
\usepackage{amsmath} 
\usepackage{mathtools} 
\usepackage{amsfonts} 
\usepackage{amssymb}
\usepackage[normalem]{ulem}

\usepackage{array}
\usepackage{booktabs}
\usepackage{rotating}
\usepackage{multirow}
\usepackage{multicol}
\usepackage{lscape}
\usepackage{subfig}
\usepackage{wrapfig}
\usepackage{comment}
\usepackage[export]{adjustbox}

\usepackage{algorithm,algorithmicx,algpseudocode}
\usepackage{color,soul}

\definecolor{babyblue}{rgb}{0.54, 0.81, 0.94}
\definecolor{armygreen}{rgb}{0.29, 0.33, 0.13}
\definecolor{brightlavender}{rgb}{0.75, 0.58, 0.89}
\definecolor{aqua}{rgb}{0.0, 1.0, 1.0}
\definecolor{caribbeangreen}{rgb}{0.0, 0.8, 0.6}
\definecolor{reddish}{rgb}{0.82, 0.1, 0.26}
\definecolor{emerald}{rgb}{0.31, 0.78, 0.47}
\definecolor{jasper}{rgb}{0.84, 0.23, 0.24}
\definecolor{red}{rgb}{1.0, 0.0, 0.0}
\definecolor{green}{rgb}{0.0, 1.0, 0.0}
\definecolor{blue}{rgb}{0.0, 0.0, 1.0}
\definecolor{darkgreen}{rgb}{0.1, 0.7, 0.1}
\definecolor{darkblue}{rgb}{0.1, 0.1, 0.7}
\definecolor{red}{rgb}{0.7, 0.1, 0.1}

\usepackage[colorlinks=true,linkcolor=blue,citecolor=blue]{hyperref}
\usepackage[colorinlistoftodos,prependcaption,textsize=tiny]{todonotes}

\usepackage{tikz,xcolor,hyperref}

\definecolor{lime}{HTML}{A6CE39}
\DeclareRobustCommand{\orcidicon}{
	\begin{tikzpicture}
	\draw[lime, fill=lime] (0,0) 
	circle [radius=0.16] 
	node[white] {{\fontfamily{qag}\selectfont \tiny ID}};
	\draw[white, fill=white] (-0.0625,0.095) 
	circle [radius=0.007];
	\end{tikzpicture}
	\hspace{-2mm}
}

\foreach \x in {A, ..., Z}{\expandafter\xdef\csname orcid\x\endcsname{\noexpand\href{https://orcid.org/\csname orcidauthor\x\endcsname}
			{\noexpand\orcidicon}}
}
			
\definecolor{darkgreen}{rgb}{0.53, 0.66, 0.42}
\definecolor{byzantine}{rgb}{0.74, 0.2, 0.64}

\begin{document}

\title{Supervised Multi-topology Network Cross-diffusion for Population-driven Brain Network Atlas Estimation}

\titlerunning{Short Title}  

\author{Islem Mhiri\index{Mhiri, Islem}\inst{1,2} \and Mohamed Ali Mahjoub\index{Mahjoub, Mohamed Ali}\inst{2} \and Islem Rekik\index{Rekik, Islem}\orcidA{}\inst{1}\thanks{ {corresponding author: irekik@itu.edu.tr, \url{http://basira-lab.com}}. This work is accepted for publication at MICCAI 2020.} }

\authorrunning{I Mhiri et al.}

\institute{$^{1}$ BASIRA Lab, Faculty of Computer and Informatics, Istanbul Technical University, Istanbul, Turkey \\ $^{2}$ Universit\'e de Sousse, Ecole Nationale d'Ing\'enieurs de Sousse, LATIS- Laboratory of Advanced Technology and Intelligent Systems, 4023, Sousse, Tunisie}

\maketitle              

\begin{abstract}
Estimating a representative and discriminative brain network atlas (BNA) is a nascent research field with untapped potentials in mapping a population of brain networks in health and disease.  Although limited, existing BNA estimation methods have several limitations. \emph{First}, they primarily rely on a similarity network diffusion and fusion technique, which only considers node degree as a topological measure in the cross-network diffusion process, thereby overlooking rich topological measures of the brain network (e.g., centrality). \emph{Second}, both diffusion and fusion techniques are implemented in fully unsupervised manner, which might decrease the discriminative power of the estimated BNAs. To fill these gaps, we propose a supervised multi-topology network cross-diffusion (SM-netFusion) framework for estimating a BNA satisfying : (i) well-representativeness (captures shared traits across subjects), (ii) well-centeredness (optimally close to all subjects), and (iii) high discriminativeness (can easily and efficiently identify discriminative brain connections that distinguish between two populations). For a specific class, given the cluster labels of the training data, we \emph{learn} a weighted combination of the topological diffusion kernels derived from degree, closeness and eigenvector centrality measures \emph{in a supervised manner}. Specifically, we learn the cross-diffusion process by normalizing the training brain networks using the learned diffusion kernels. This normalization well captures shared networks between individuals at different topological scales, improving the representativeness and centeredness of the estimated multi-topology BNA. Our SM-netFusion produces the most centered and representative template in comparison with its variants and state-of-the-art methods and further boosted the classification of autistic subjects by 5 to 15\%. SM-netFusion presents the first work for supervised network cross-diffusion based on graph topological measures, which can be further leveraged to design an efficient graph feature selection method for training predictive learners in network neuroscience. Our SM-netFusion code is available at \url{https://github.com/basiralab/SM-netFusion}.
\end{abstract}

\keywords{Brain network atlas learning $\cdot$ Supervised network cross-diffusion and fusion $\cdot$ Heterogeneous manifold learning}

\section{Introduction}

Estimating a representative and discriminative \emph{brain network atlas} (BNA) marked a new era for mapping a population of brain networks in health and disease. A few recent landmark studies have relied on developing the concept of a network atlas estimated from a population of brain networks. One pioneering work includes \cite{rekik2017} on estimating a brain network atlas from a population of both morphological and functional brain networks using diffusive-shrinking graph technique \cite{wang2014}. Later, \cite{rekik2018b} introduced brain the morpho-kinectome (i.e., population-based brain network atlas) to investigate the relationship between brain morphology and connectivity kinetics in developing infants. Another work \cite{dhifallah2018} proposed the concept of population-driven connectional brain template for multi-view brain networks using a cluster-based diffusion and fusion technique. More recently, \cite{dhifallah2020} designed a sample selection technique followed up by a graph diffusion and fusion step. \cite{mhiri2020} estimated also a brain network atlas-guided feature selection (NAGFS) method to differentiate the healthy from the disordered connectome. 

Although they presented compelling results, these works have several limitations. \emph{First}, all these promising works \cite{rekik2017,rekik2018b,dhifallah2018,dhifallah2020,mhiri2020} have relied on the similarity network fusion (SNF) and diffusion technique introduced in \cite{wang2014}. Although compelling, \cite{wang2014} non-linearly diffuses and fuses brain networks without considering their heterogeneous distributions or the possibility of them lying on different subspaces. This might not preserve the pairwise associations between different networks in complex manifold they sit on. \emph{Second}, \cite{wang2014} \emph{solely} uses node degree as a topological measure in the cross-network diffusion process. However, measures of the degree or strength provide only partial information of the role (significance) of a node in a network. So, one cannot capture the full structure of a network because node degree only considers the immediate and local neighborhood of a given node (i.e., anatomical region of interest (ROI) in a brain network). Also, it treats all node connections equally \cite{fornito2016}. Hence, it captures the quantitative aspect of node (how many neighbors it has) but not the qualitative aspect of a node (the quality of its neighbors). \emph{Third}, both diffusion and fusion techniques were implemented in a fully \emph{unsupervised} manner without considering the heterogeneous distribution of the brain network population (e.g., typical or autistic), which would eventually affect the representativeness of the estimated BNAs. 

To address all these limitations, we propose a supervised multi-topology network cross-diffusion (SM-netFusion) framework for \emph{learning} a BNA which satisfies the following constraints: (i) it is \emph{well-representative} that consistently captures the unique and distinctive traits of a population of functional networks, (ii) it is \emph{well-centered} that  occupies a center position optimally near to all  individuals, and (iii) it reliably identifies the most \emph{discriminative} disordered brain connections by comparing templates estimated using disordered and healthy brains, respectively. \emph{First}, to handle data heterogeneity within each specific class, we learn the pairwise similarities between connectomes and map them into different subspaces where we assign to each brain network living in the same subspace the same label. This clustering step allows to explore the underlying data distribution prior to the diffusion process for BNA estimation. \emph{Second}, for each training sample in the given class, we define a tensor stacking as frontal views the degree, closeness and eigenvector centrality matrices. By fusing the tensor frontal views, we generate an average topological matrix which nicely characterize both local and global relationships between brain ROIs. \emph{Third}, to preserve the heterogeneous distribution of the data in a specific class, we supervisedly learn a \emph{subject-specific weight} to map each average topological matrix to its cluster label. Next, for each subject, we multiply each weight with its training average topological matrix to generate \emph{the normalization kernel}, whose inverse normalizes the original brain network. \emph{Fourth}, in a specific class, we nonlinearly cross-diffuse the normalized brain networks so that all diffused networks lie close to each other for the final fusion step to generate the target of a specific class. The proposed cross-diffusion process well captures shared connections across individuals at different topological scales, improving the representativeness and centeredness of the estimated multi-topology BNA. More importantly, by \emph{comparing} the learned healthy and disordered BNAs, we also investigate the discriminative potential of our estimated brain network atlases in reliably differentiating between typical and disordered brains which can be eventually used to train a predictive learner for accurate and fast diagnosis. The main contributions of our method are three-fold:

\begin{enumerate}

\item \emph{On a methodological level.} SM-netFusion presents the first work on \emph{supervised} and \emph{class-specific} network cross-diffusion based on graph topological measures, which can be also leveraged to design an efficient feature selection method for training predictive learners in network neuroscience.

\item \emph{On a clinical level.}  By comparing BNAs produced by SM-netFusion in healthy and disordered groups, one can easily spot a connectional fingerprint of a disorder (i.e., a set of altered brain connectivities).

\item \emph{On a generic level.} Our framework is a generic method as it can be applied to brain networks derived from any neuroimaging modality (e.g., morphological and structural connectomes) given that they are isomorphic.
\end{enumerate}

\begin{figure}[htpb!]
\centering
\vspace{-10pt}
\includegraphics[width=12.5cm]{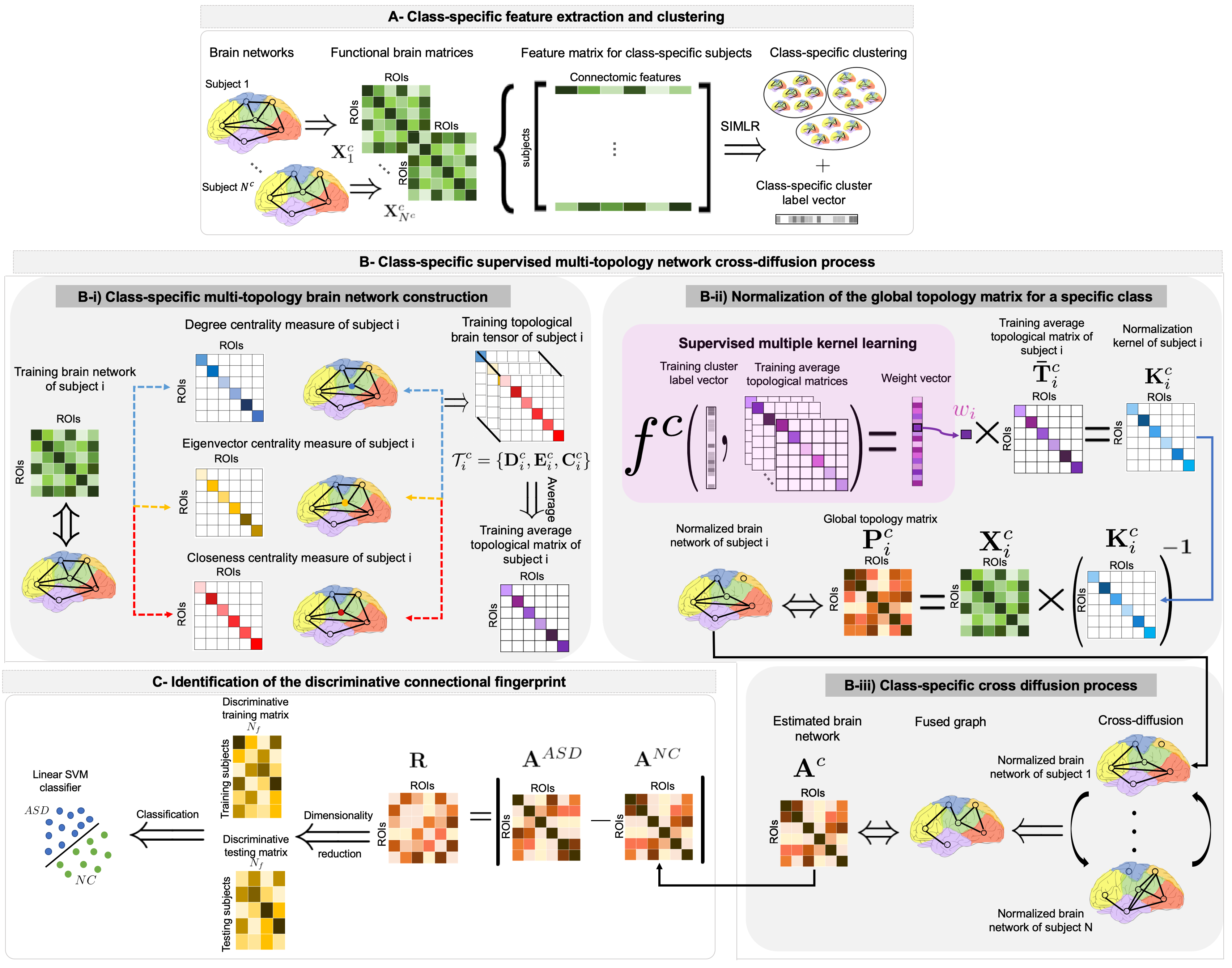}
\caption{\emph{Illustration of the proposed supervised multi-topology network cross-diffusion (SM-netFusion) framework with application to brain connectomes.} \textbf{(A) Class-specific feature extraction and clustering}. For each subject $i$ in class $c$, we vectorize the upper triangular part of its connectivity matrix $\mathbf{X}^c_{i}$. Next, we concatenate all feature vectors into a data feature matrix which we cluster similar functional brain networks into non-overlapping subspaces using SIMLR framework \cite{wang2018} where we assign to each brain network living in the same subspace the same label. \textbf{(B) Class-specific supervised multi-topology network cross-diffusion.} For each training sample $i$ in class $c$, we define a tensor $\mathcal{T}^c_i$ stacking as frontal views the degree, closeness and eigenvector centrality matrices. By fusing the tensor views,we generate an average topological matrix $\mathbf{\bar{T}}^c_i$. Next, to preserve the heterogeneous distribution of the data in class $c$, we supervisedly learn a subject-specific weight to map each $\mathbf{\bar{T}}^c_i$ to its cluster label. Then, for each subject, we multiply its learned weight with its $\mathbf{\bar{T}}^c_i$ to generate the normalization kernel $\mathbf{K}^c_{i}$, whose inverse normalizes the original brain network. Later, we nonlinearly cross-diffuse the normalized brain networks so that all diffused networks lie close to each other for the final fusion step to generate the target BNA.
 \textbf{(C) Identification of the discriminative connectional biomarker.} By computing the absolute difference matrix $\mathbf{R}$ between $\mathbf{A}_{ASD}$ and $\mathbf{A}_{NC}$ network atlases, we select the top $N_f$ features with the highest discrepancy and use those to train a linear support vector machine (SVM) classifier within a five-fold cross-validation scheme. }
\label{fig:1}
\end{figure}

\section{Proposed Method}
In the following, we present the main steps of the proposed SM-netFusion framework for estimating a representative, centered and discriminative BNA (\textbf{Fig.}~\ref{fig:1}).

\textbf{A- Class-specific feature extraction and clustering}. Given a population of $N^c$ brain networks in class $c$, each network $i$ is encoded in a symmetric matrix $\mathbf{X}_{i}^c \in \mathbb{R}^{r \times r}$, where $r$ denotes the number of anatomical regions of interest (ROIs). Since each matrix $\mathbf{X}_{i}^c$ is symmetric, we extract a feature vector for subject $i$ in class $c$ by simply vectorizing its upper off-diagonal triangular part. Next, we horizontally stack feature vectors of all subjects to define a data feature matrix of size $N^c \times \frac{r \times (r-1)}{2}$ (\textbf{Fig.}~\ref{fig:1}-A). Next, we disentangle the heterogeneous distribution of the brain networks by clustering similar functional brain networks into non-overlapping subspaces using Single Cell Interpretation via Multikernel Learning (SIMLR) framework \cite{wang2018} where we assign to each brain network living in the same subspace the same cluster label. This clustering step allows to explore the underlying data distribution prior to the diffusion process for the BNA estimation.

\textbf{B- Class-specific supervised multi-topology network cross-diffusion}. \textbf{Fig.}~\ref{fig:1}-B illustrates the key steps of the proposed SM-netFusion framework, which we detail below. 

\emph{i- Class-specific multi-topology brain network construction.} In the first step, for each training subject in class $c$, we compute the most commonly used centrality measures in brain networks (degree centrality, eigenvector centrality and closeness centrality) \cite{fornito2016} (\textbf{Fig.}~\ref{fig:1}-B- i). These topological measures define the central nodes where each communication in the network should pass through them. \cite{freeman1979} reported that three fundamental properties can be ascribed to the central node: (1) It has the maximum degree because it is connected to all other nodes. (2) It is  the best mediator that belongs to the shortest path
between all pairs of nodes. (3) It is maximally close to all other nodes. Particularly, the degree centrality measures the number of edges connecting to a node (ROI). The \emph{degree centrality} $D(n)$ of a node $n$ is defined as: $D(n) = \sum_{n\neq k} \mathbf{A}_{nk}$, where $\mathbf{A}_{nk}=1$ if the connectivity of node $n$ and node $k$ exists; otherwise $\mathbf{A}_{nk}=0$.
 The degree centrality defines the central nodes with the highest number of degree or connections. It examines the immediate neighbors of the node. So, in our case, it characterizes the local topology of each brain region. \emph{Eigenvector centrality} is the first eigenvector of the  brain connectivity matrix, which corresponds to the largest eigenvalue $\lambda_1$ (called the principal eigenvalue): $E(n) = \frac{1}{\lambda_{1}}\times \sum^{r}_{k=1} \mathbf{A}_{nk}\mathbf{x}_{k}$, $\mathbf{A}_{nk}$ is the connectivity strengths between nodes $n$ and $k$, and $\mathbf{x}$ is a nonzero vector that, when multiplied by $\mathbf{A}$, satisfies the condition $\mathbf{A}\mathbf{x} = \lambda \mathbf{x}$. The \emph{closeness centrality} $C(n)$ reflects the closeness between a node $n$ and other nodes in a brain network: $C(n) = \frac{r-1}{\sum_{n \neq k} {l}_{nk}}$, where $l_{nk}$ is the shortest path length between nodes $n$ and $k$. This centrality measure defines the mean distance between the central node and all other nodes in a network. It captures the effective outreach via closest path. Specifically, the node with the highest closeness will affect all other nodes in a short period of time (shortest path). These topological measures have been extensively examined in the literature of network neuroscience, where brain function integration and segregation was shown to work through brain hubs (i.e., central nodes) \cite{Bassett2017}. In fact, these centrality measures define the most significant ROIs (central nodes) nesting function and cognitive neural flow. Hence, we adopt these metrics in order to characterize both local and global relationships between brain ROIs. Once the topological matrices are defined for each subject, we stack them into a tensor $\mathcal{T}^c_i=\{\mathbf{D}^c_i,\mathbf{E}^c_i,\mathbf{C}^c_i\}$, which is fused into an average topological matrix $\mathbf{\bar{T}}^c_i$ in class $c$ (\textbf{Fig.}~\ref{fig:1}-B- i). 

\emph{ii- Supervised multiple kernel normalization.} To preserve the heterogeneous distribution of the data in class $c$, we supervisedly learn a subject-specific weight to map each average topological matrix to its cluster label. Hence, we apply a supervised machine learning method based on multiple kernel learning (MKL) called EasyMKL \cite{aiolli2015} in order to find the optimal mixture of kernels over the different training average topological matrices (\textbf{Fig.}~\ref{fig:1}-B- ii). EasyMKL achieves higher scalability with respect to the number of kernels to be combined at a low computational cost in comparison with other MKL methods. Given a class $c$, we learn a mapping $f^c$ (i.e., a weight vector $\mathbf{w} \in \mathbb{R}^{n_{tr} \times 1}$) transforming the training average topological matrices (i.e., a set of kernels) onto their corresponding cluster labels by solving a simple quadratic optimization loss:

\begin{gather}
    f^c= \min_{\mathbf{w}:||\mathbf{w}||_{2}=1} \min_{\mathbf{\gamma} \in \Gamma} \mathbf{\gamma}^{T} \mathbf{Y}  (\sum_{i=0}^{n_{tr}} w_{i} \mathbf{\bar{T}}^c_i) \mathbf{Y}\mathbf{\gamma} + \lambda ||\mathbf{\gamma}||^{2},
\end{gather}

where $\mathbf{Y}$ is a diagonal matrix with training cluster labels on the diagonal, $\lambda$ is a regularization hyper-parameter, $\textbf{w}$ is the weight vector that maximizes the pairwise margin between different subspaces. The domain $\Gamma$ represents the domain of probability
distributions $\mathbf{\gamma} \in \mathbb{R}_{+}^{n_{tr}}$ defined over the sets of subspaces (clusters), that is ${\Gamma}=\left\{\mathbf{\gamma}\in \mathbb{R}_{+}^{n_{tr}} | \sum_{i \in subspace_j} \mathbf{\gamma}_{i}=1, j=1, \dots, n_c\right\}$. It turns out this quadratic functional has a closed form solution where 
each learned weight coefficient for subject $i$ is defined as $w_{i}=\mathbf{\gamma}^{T} \mathbf{Y}(\frac{\mathbf{\bar{T}}^c_i}{Tr(\mathbf{\bar{T}}^c_i)}) \mathbf{Y}\mathbf{\gamma} $, where $Tr(\mathbf{\bar{T}}^c_i)$ is the trace of a basic kernel (i.e., average topological matrix of subject $i$). Next, we multiply each weight with its training average topological matrix to generate the normalization kernel $\mathbf{K}_i^c$ of each subject in class $c$.

\emph{iii- Class-specific cross-diffusion process for BNA estimation.} Previously, to learn the cross-diffusion process in the SNF \cite{wang2014} technique, one needs to first define a status matrix, also referred to as the global topology matrix $\mathbf{P}_i$, capturing the global structure of each individual $i$ and carrying the full information about the similarity of each ROI to all other ROIs. This status matrix is iteratively updated by diffusing its structure across the average global structure of other brain networks. Conventionally, $\mathbf{P}_i$ is a normalized weight matrix $\mathbf{P}_i(k,l) = \begin{cases} \frac{1}{2}\mathbf{D}_i^{-1}\mathbf{X}_i(k,l) &  l \neq k \\1/2, \ l = k \\ \end{cases}$, where $\mathbf{D}_i$ is the diagonal degree (strength) matrix of subject $i$ \cite{wang2014,rekik2018b,dhifallah2018,dhifallah2020,mhiri2020}. However, this normalization overlooks  rich  topological  measures  of  the  brain  network  (e.g.,  centrality) since the degree measure only focuses on the immediate and local neighborhood of a node. One way of casting a more \emph{topology-perserving normalization} of a brain network $i$ in class $c$ is by using the learned normalization kernel $\mathbf{K}_i^c$ to define a \emph{multi-topology aware} status matrix as: $\mathbf{P}^c_i(k,l) = \begin{cases} \frac{1}{2}(\mathbf{K}^c_i)^{-1}\mathbf{X}^c_i(k,l) &  l \neq k \\1/2, \ l = k \\ \end{cases}$, where $\mathbf{X}^c_i(k,l)$ denotes the connectivity between ROIs $k$ and $l$ (\textbf{Fig.}~\ref{fig:1}--B-ii). 
Next, for class $c$, we define a kernel similarity matrix $\mathbf{Q}^c_i$ for each individual $i$, which encodes its local structure by computing the similarity between each of its elements ROI $k$ and its nearest ROIs $l$ as follows: $\mathbf{Q}^c_i(k,l) =
\begin{cases}
	\frac{\mathbf{X}^c_i(k,l)}{\sum_{p \in n_{k}} \mathbf{X}^c_i(k,p)} & l \in n_{k} \\
	0, &  otherwise \\
\end{cases}$, where $n_{k}$ represents the set of $q$ neighbors of ROI $k$ identified using $KNN$ algorithm. In order to integrate the different networks into a single network, each multi-topology matrices $\mathbf{P}^c_i$ is iteratively updated for each individual by diffusing the topological structure of $\mathbf{P}^c_j$ of $N^c-1$ networks ($j \neq i$) along the local structure $\mathbf{Q}^c_i$ of subject $i$ as follows:  $\mathbf{P}^c_i = \mathbf{Q}^c_i \times \bigg(  \frac{\sum_{j \neq i} \mathbf{P}^c_j }{ N^c - 1} \bigg)  \times  (\mathbf{Q}^c_i)^T, \  j \in \{ 1, \dots, N^c \} $, where $\frac{\sum_{j \neq i} \mathbf{P}^c_j }{N^c - 1}$ denotes the average diffused networks in class $c$ excluding subject $i$. This step is iterated $n^{\star}$ times and generates $N^c$ parallel interchanging diffusion processes on $N^c$ networks. If two connectivities are similar in all data types, their similarity will be enhanced through the diffusion process and vice versa. By fusing the cross-diffused networks within class $c$, we estimate the target BNA: $\mathbf{A}^c = \frac{\sum_{i=1}^{N^c} (\mathbf{P}^c_i)^{n^{\star}}}{N^c}$ (\textbf{Fig.}~\ref{fig:1}-B- iii).

\textbf{C- Identification of the discriminative connectional fingerprint.} To investigate the discriminative power of our estimated brain network atlas, we  select the most relevant features distinguishing between two populations by computing the absolute difference between both estimated training network atlas matrices $\mathbf{A}^{ASD}$ and $\mathbf{A}^{NC}$ as follows: $ \mathbf{R}(NC, ASD)= |\mathbf{A}^{ASD} - \mathbf{A}^{NC}|$ (\textbf{Fig.}~\ref{fig:1}-C). By taking all elements in the upper off-diagonal part of the residual matrix $\mathbf{R}$, we select the top $N_f$ features with the largest non-zero values as these identify the brain connectivities where both BNAs largely differ. Next, using the top $N_f$ selected connectivities derived from the training set, we train a linear support vector machine (SVM) classifier. In the testing stage, we extract the same features from the testing functional networks, then pass the selected features to the trained classifier for predicting the labels of the testing subjects.

\begin{figure}[!htpb]
\centering
\vspace{-10pt}
\includegraphics[width=12cm]{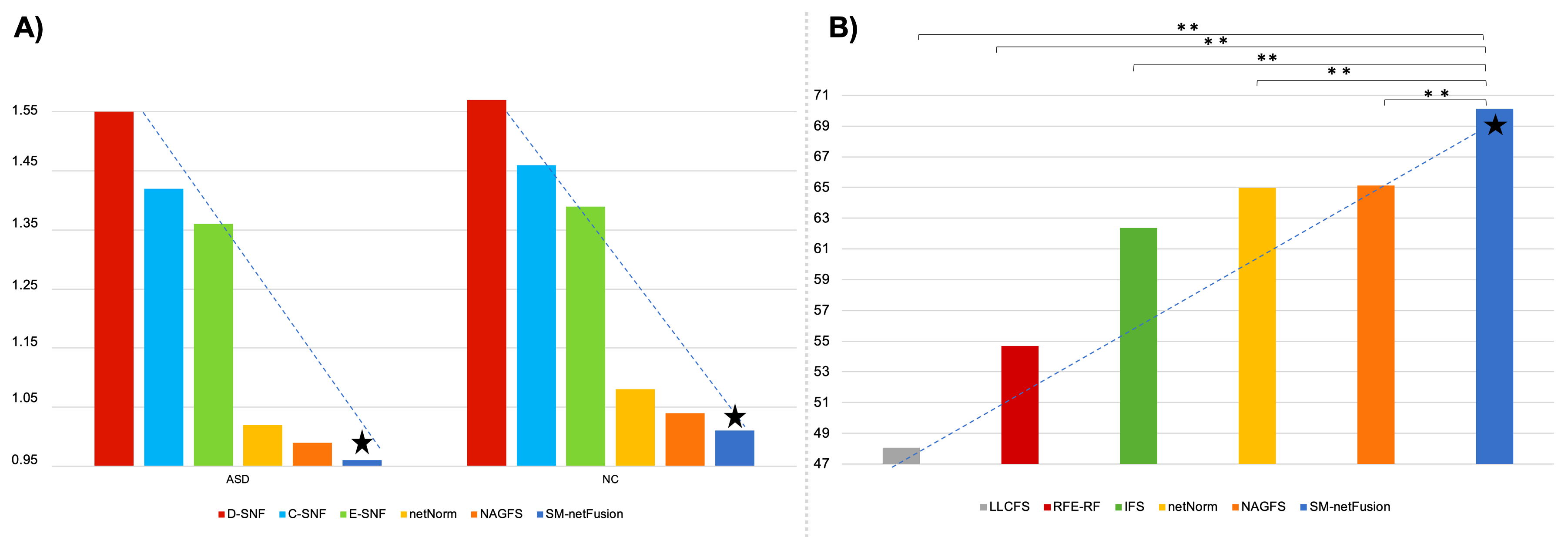}
\caption{\emph{Evaluation}. \textbf{A) Evaluation of the estimated network atlas for NC and ASD populations using different fusion strategies.} We display the mean Frobenius distance between estimated brain network atlas and all individual networks in the population using D-SNF \cite{wang2014}, C-SNF, E-SNF, netNorm \cite{dhifallah2020}, NAGFS \cite{mhiri2020} and SM-netFusion. Clearly, SM-netFusion achieves the minimum distance in both ASD and NC groups. \textbf{B) ASD/NC classification using different brain networks}. Average classification accuracies for our method (SM-netFusion+SVM), (IFS+SVM) \cite{roffo2015}, (netNorm+SVM) \cite{dhifallah2020}, (RF-RFE) \cite{nembrini2013} and (LLCFS+SVM) \cite{zeng2010}. The best performance was achieved by our method. $\star$: Our method and $(**)$ for p-value $< 0.05$ using two-tailed paired t-test. }
\label{fig:2}
\end{figure}

\section{Results and Discussion}
\textbf{Evaluation dataset and parameters.} We used five-fold cross-validation to evaluate the proposed SM-netFusion framework on 505 subjects (266 ASD and 239 NC) from Autism Brain Imaging Data Exchange (ABIDE) preprocessed public dataset \footnote{\url{http://preprocessed-connectomes-project.org/abide/}}. Several preprocessing steps were implemented by the data processing assistant for resting-state fMRI (DPARSF) pipeline. Each brain rfMRI was partitioned into 116 ROIs. For SIMLR parameters \cite{wang2018}, we tested SM-netFusion using $n_c=\{ 1, 2, \dots, 6\}$ clusters and we found that the best result was $n_c=3$. For the cross-diffusion process parameters, we also set  the number of iterations $n^{\star} = 20$ as recommended in \cite{wang2014} for convergence. We fixed the number of closest neighbors $K = 25$ across comparison methods.

\textbf{Evaluation and comparison methods.} \emph{Representativeness.} To evaluate the centeredness and representativeness of our brain network atlas estimation, we benchmarked our method against five network fusion strategies: (1) D-SNF method \cite{wang2014} which considers only node degree as a topological measure in the cross-network diffusion process, (2) C-SNF method which considers only closeness centrality as a topological measure, (3) E-SNF method which considers only eigenvector centrality as a topological measure, (4) netNorm\footnote{\url{https://github.com/basiralab/netNorm-PY}} method \cite{dhifallah2020} which uses a high-order sample selection technique to build a connectional template, and (5) NAGFS\footnote{\url{https://github.com/basiralab/NAGFS-PY}} method which uses SNF diffusion and fusion techniques along with clustering. 

\begin{wrapfigure}{r}{0.5\textwidth}
  \begin{center}
	\vspace{-10pt}
    \includegraphics[width=0.48\textwidth]{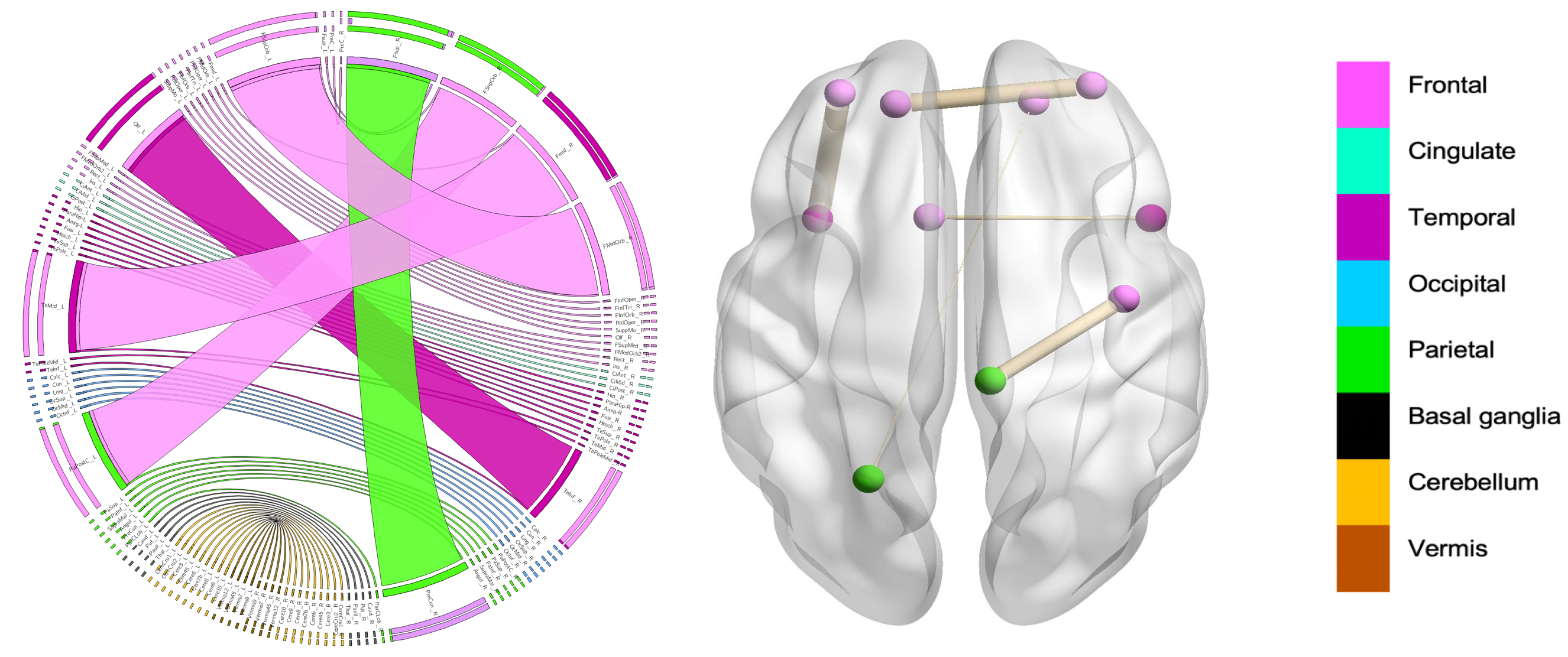}
  \end{center}
	\vspace{-10pt}
  \caption{\emph{The strongest connections present the 5 most discriminative network connections
between ASD and NC groups}. The circular graphs were generated using Circos table viewer \cite{Krzywinski2009}. We used BrainNet Viewer Software \cite{xia2013} to display the regions of interest involving the most discriminative connectivities.}
\label{fig:4}
\end{wrapfigure}

As illustrated in \textbf{Fig.}~\ref{fig:2}-A, we computed the mean Frobenius distance defined as as $d_F (\mathbf{A},\mathbf{B})= \sqrt{\sum_{i}\sum_{j}| a_{ij}-b_{ij}|^{2}}$ between the estimated network atlas and individual networks in the population. A smaller distance indicates a more centered network atlas with respect to all individuals in the population. We observe that our proposed multi-topology BNA remarkably outperforms conventional techniques by achieving the minimum distance for both ASD and NC populations.

\emph{Discriminativeness}. Furthermore, we demonstrate that SM-netFusion produces highly discriminative BNAs in terms of identifying the most discriminative brain connections between two classes. Specifically, we conducted a comparative study between ASD and NC populations using the estimated multi-topology BNAs. Using 5-fold cross-validation strategy, we trained an SVM classifier using the top $N_f$ most discriminative features identified by each of the following feature
selection methods: (1) recursive feature elimination with random forest (RFE-RF) \cite{granitto2006}, (2) local learning-based clustering feature selection (LLCFS) \cite{zeng2010}, (3) infinite feature selection (IFS) \cite{roffo2015}, (4) netNorm \cite{dhifallah2020}, and (5) a  NAGFS \cite{mhiri2020}). Clearly, our method significantly outperformed all comparison methods results as shown in \textbf{Fig.}~\ref{fig:2}-B in terms of classification accuracy ($p-value< 0.05$ using two-tailed paired t-test). Our results also demonstrate
that our SM-netFusion for BNA estimation outperforms state-of-the-art methods along with its ablated versions in terms of both representativeness and discriminativeness. 

\emph{Neuro-biomarkers}.  \textbf{Fig.}~\ref{fig:4} displays the top 5 discriminative ROIs distinguishing between healthy and autistic subjects by computing the absolute difference between the estimated BNAs and pinning down regions with highest differences. We notice that most of the discriminative functional brain connectivities involved the frontal lobe. Indeed, previous studies reported that the frontal lobe has a major role in speech and language production, understanding and reacting to others, forming memories and making decisions which might explain the prevalence of altered brain connectivities in this brain lobar region \cite{kumar2010}. Our SM-netFusion is a generic framework for supervised graph integration, which can be further leveraged to design an efficient graph feature selection method for training predictive learners for examining graph-based data representations.

\section{Conclusion}
We proposed the first work for supervised network cross-diffusion based on graph topological measures (SM-netFusion) by enhancing the non-linear fusion process using a weighted mixture of multi-topological measures.  Our framework can be also leveraged to design an efficient feature selection method for training predictive learners in network neuroscience. The proposed SM-netFusion produces the most centered and representative BNAs in comparison with its variants as well as state-of-the-art methods and further boosted the classification of autistic subjects by $5-15\%$. In our future work, we will evaluate our framework on larger connectomic datasets covering a diverse range of neurological disorders such as brain dementia. Furthermore, we aim to explore the discriminative power of brain network atlases derived from other brain modalities such as structural \cite{park2013structural} and morphological brain networks \cite{nebli2019gender,bilgen2020machine}.

\section{Supplementary material}

We provide three supplementary items on SM-netFusion for reproducible and open science:

\begin{enumerate}
	\item A 5-mn YouTube video explaining how SM-netFusion works on BASIRA YouTube channel at \url{https://youtu.be/eWz65SyR-eM}.
	\item SM-netFusion code in Matlab on GitHub at \url{https://github.com/basiralab/SM-netFusion}. 
	\item SM-netFusion code in Python on GitHub at \url{https://github.com/basiralab/SM-netFusion-PY}. 
\end{enumerate}


\section{Acknowledgments}

I. Rekik is supported by the European Union's Horizon 2020 research and innovation programme under the Marie Sklodowska-Curie Individual Fellowship grant agreement No 101003403 (\url{http://basira-lab.com/normnets/}).

\bibliography{Biblio3}
\bibliographystyle{splncs}
\end{document}